\documentclass[10pt,twocolumn,letterpaper]{article}

\usepackage[pagenumbers]{cvpr} 
\usepackage{times}
\usepackage{epsfig}
\usepackage{graphicx}
\usepackage{amsmath}
\usepackage{amssymb}
\usepackage{svg}
\usepackage{booktabs}
\usepackage{multirow}

\begin{document}

\title{Database-Agnostic Gait Enrollment using SetTransformers}

\author{Nicoleta Basoc, Adrian Cosma, Andy Cǎtrunǎ, Emilian Rǎdoi\\
National University of Science and Technology POLITEHNICA Bucharest\\
{\tt\small nicoleta\_nina.basoc@stud.acs.upb.ro} \\ 
{\tt\small\{ioan\_adrian.cosma, andy\_eduard.catruna, emilian.radoi\}@upb.ro}
}

\maketitle
\thispagestyle{empty}

\begin{abstract}
Gait recognition has emerged as a powerful tool for unobtrusive and long-range identity analysis, with growing relevance in surveillance and monitoring applications. Although recent advances in deep learning and large-scale datasets have enabled highly accurate recognition under closed-set conditions, real-world deployment demands open-set gait enrollment, which means determining whether a new gait sample corresponds to a known identity or represents a previously unseen individual. In this work, we introduce a transformer-based framework for open-set gait enrollment that is both dataset-agnostic and recognition-architecture-agnostic. Our method leverages a SetTransformer to make enrollment decisions based on the embedding of a probe sample and a context set drawn from the gallery, without requiring task-specific thresholds or retraining for new environments. By decoupling enrollment from the main recognition pipeline, our model is generalized across different datasets, gallery sizes, and identity distributions. We propose an evaluation protocol that uses existing datasets in different ratios of identities and walks per identity. We instantiate our method using skeleton-based gait representations and evaluate it on two benchmark datasets (CASIA-B and PsyMo), using embeddings from three state-of-the-art recognition models (GaitGraph, GaitFormer, and GaitPT). We show that our method is flexible, is able to accurately perform enrollment in different scenarios, and scales better with data compared to traditional approaches. We will make the code and dataset scenarios publicly available.
\end{abstract}


\section{Introduction}
\label{sec:intro}
Analyzing human walking patterns has emerged as a robust method for automatic monitoring due to its non-invasive nature and effectiveness from a distance without requiring subject cooperation \cite{bashir2010gait,cosma2023exploring}. Gait-based systems leverage the uniqueness of walking styles to identify people and analyze their attributes \cite{cosma2022learning}, proving particularly useful for surveillance and monitoring applications \cite{parashar2023real}.

Training robust gait-based recognition systems has become feasible due to advances in deep learning methods \cite{vaswani2017attention, dosovitskiy2020image, liu2021swin} and the emergence of large-scale datasets collected from public surveillance cameras \cite{cosma2021wildgait} that can be annotated manually \cite{zhu2021gait} or automatically \cite{cosma2022learning}. However, a relatively unexplored challenge arises for the practical deployment of such systems: determining whether a new gait sample belongs to a person already enrolled in the database (present in the gallery) or represents a new person that should be enrolled. This scenario, commonly referred to as open-set recognition or enrollment \cite{mazzieri2025open,ni2021open} differs from the scenario explored by most works on gait analysis and recognition, which consists of operating in a closed set setting with all testing identities` enrolled in the gallery. 

The gait enrollment problem is highly relevant for the real-world deployment of gait recognition models. A possible real-world enrollment scenario would involve explicit subject cooperation, such as subjects scanning an ID card while having their walk enrolled in the database. If this is not the case, operators manually annotate new identities, which is inefficient and time-consuming. Despite its relevance to real-world gait applications, open-set gait recognition remains relatively underexplored, with most works primarily focusing on closed-set scenarios. Most methods on open-set gait recognition utilize threshold-based approaches \cite{ni2021open, mazzieri2025open}, which require manual tuning and do not generalize to new databases containing different identities from those in the training set. 

We propose a novel method for solving the enrollment problem in gait recognition: we train a SetTransformer \cite{lee2019set} model, which makes the enrollment decision based on the target embedding and contextual information in the form of potentially relevant embeddings. In contrast to other enrollment approaches \cite{yang2022multiscenario, ni2021open, mazzieri2025open}, we decouple our model from the gait recognition training pipeline and train it in a \textit{dataset-agnostic} fashion by presenting it with samples that come from different gallery-probe settings. This ensures that the model does not overfit to a single database configuration, but generalizes to multiple scenarios. Our model is decoupled from the main recognition architecture, operating only on its generated embeddings. This makes our method agnostic of the recognition architecture and the gait modality, offering great flexibility in deployment across different systems. We demonstrate the effectiveness of our approach using skeleton data in conjunction with three popular gait recognition models: GaitGraph \cite{teepe2021gaitgraph}, GaitFormer \cite{cosma2022learning}, and GaitPT \cite{catruna2024gaitpt} and two controlled gait recognition datasets: CASIA-B \cite{yu2006framework} and PsyMo \cite{cosma2023psymo}.

This work makes the following contributions:
\begin{itemize}
    \item We propose a gait enrollment setup designed to train and evaluate a model that determines whether new gait samples belong to known identities in the gallery or represent previously unseen individuals. Our setup automatically builds different enrollment scenarios from a gait recognition dataset, varying the number of identities in the gallery and the number of samples per identity for more comprehensive training and evaluation.
    
    \item We propose a flexible SetTransformer-based model that predicts the enrollment status of new gait samples based on gait embeddings and selected contextual information from the gallery. Due to its training and input, the model is \textbf{dataset-agnostic}, by training on different configurations of gallery-probe sets, and \textbf{model-agnostic}, by not being tied to a particular gait recognition model training pipeline.
\end{itemize} 

\section{Related Work}
\label{sec:related}
Most works on vision-based gait recognition \cite{chao2019gaitset, fan2020gaitpart, lin2022gaitgl, fan2024skeletongait} assume a closed-set setting, in which identities belonging to the probe set for evaluation also exist in the gallery. However, real-world recognition systems cannot rely on this assumption \cite{Gunther_2017_CVPR_Workshops} and should adopt the open-set identification framework. In open-set recognition, the system must determine whether a sample belongs to a known identity or represents a new individual that should be enrolled in the gallery.

The literature on open-set gait recognition remains relatively unexplored, with existing works drawing from open-set face identification \cite{sun2015deepid3, Gunther_2017_CVPR_Workshops, yu2019unknown, shu2020p, zhang2020hybrid, vaze2021open} and primarily employing radar data \cite{yang2019open, yang2022multiscenario, ni2021open, mazzieri2025open} for processing the walking patterns. A straightforward approach to determining whether a sample is unknown is using threshold-based methods, similar to other works on open-set face recognition \cite{sun2015deepid3, Gunther_2017_CVPR_Workshops, shu2020p}. For example, Ni et al. \cite{ni2021open} construct statistical models for each known identity based on intra-class distances in the embedding space,  allowing the model to estimate the probability that a new sample belongs to a specific identity, classifying it as unknown if the highest probability is below a threshold. Similarly, Mazzieri et al. \cite{mazzieri2025open} add an adversarial loss that forces embeddings to follow a Gaussian distribution for each known identity. They construct a mixture of Gaussians based on which the model estimates the probability of a new sample being known and sets a threshold for identifying unknown subjects.

A similar method is that of Yang et al. \cite{yang2022multiscenario}, who employs an ensemble of 2 detectors to predict whether a gait sample belongs to an unknown identity. One detector evaluates the VAE reconstruction error, while the other analyzes the embedding of the gait in relation to known embeddings. As shown by Gunther et al. \cite{Gunther_2017_CVPR_Workshops}, threshold-based methods are not the most effective solution to open-set recognition problems, which is why our approach utilizes a separate model that learns to classify samples based on dataset statistics.

Utilizing additional data labeled as the unknown class is a common strategy employed in open-set recognition works. To bypass the limitations of threshold-based approaches, Yang et al. \cite{yang2019open} utilize a Generative Adversarial Network to synthesize gait data, which acts as the unknown samples during training. In this setting, the discriminator is trained to classify gait samples into known identities or an extra class reserved for the unknown, synthetic examples. A similar approach from the face identification literature is that of Yu et al. \cite{yu2019unknown}, who employ unlabeled face images as samples with the unknown class. Their face recognition model is trained to predict low confidence for these samples, encouraging the classifier to utilize specific parts of the embedding space for unknown identities.

Our evaluation protocol closely resembles real-world scenarios, where the gallery can have a variable number of identities and samples per subject. This variability renders approaches that rely on a fixed gallery setting ineffective. Moreover, our model is dataset-agnostic because of its training with various enrollment scenarios, in contrast to \cite{mazzieri2025open, ni2021open}, which depends on a dataset constructed specifically for open-set recognition with specific known and unknown identities. We construct our model to be decoupled from the recognition architecture, using its embeddings as input, enabling compatibility with any gait recognition architecture and modality, unlike methods that either directly operate on the input modality \cite{yang2019open, yang2022multiscenario} or have to modify the specific recognition model \cite{mazzieri2025open}.

\section{Method}
\label{sec:method}

\subsection{Description of Gait Enrollment Setup}

\begin{figure*}
    \centering
    \includesvg[width=0.75\linewidth]{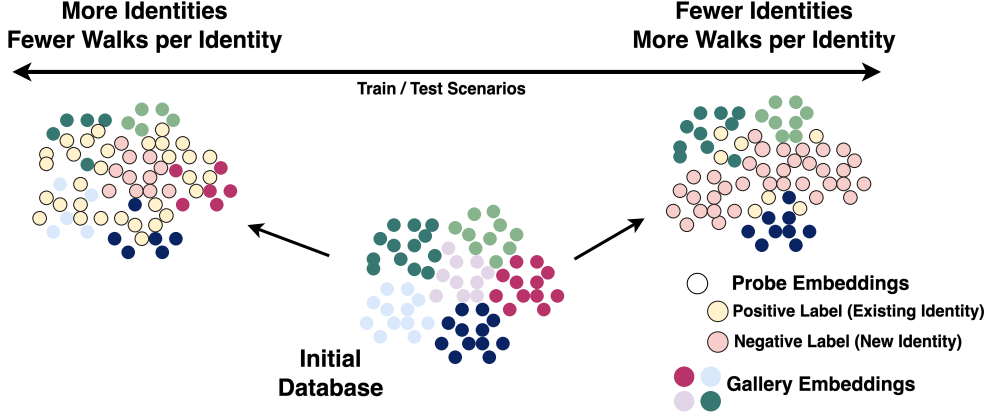}
    \caption{Overview of enrollment training and evaluation scenarios. Starting from an initial gallery of identity embeddings, probes are evaluated under database setups containing more identities but with fewer walks per identity and fewer identities containing more walks per identity. Each scenario is described with a certain id:walk ratio. Positive probes match existing gallery identities; negative probes represent new identities.}
    \label{fig:enrollment-setup}
\end{figure*}

We follow standard gait‐recognition terminology by partitioning our data into a \textit{gallery set} — a collection of enrolled identity embeddings — and a disjoint \textit{probe set} of queries. Unlike recognition, where the task is to assign each probe to one of the known identities, our objective is enrollment: for each probe embedding, the model must decide whether it belongs to one of the gallery identities or represents a novel individual.

In contrast to other works \cite{mazzieri2025open, ni2021open} that specifically build an open-set recognition dataset, we construct our training and evaluation setup for enrollment from existing datasets (CASIA-B \cite{yu2006framework} and PsyMo \cite{cosma2023psymo}). Real-world scenarios rarely have a balanced distribution of identities or number of walking instances per identity. As such, we develop multiple scenarios for training and testing the gait enrollment model by varying the id:walk ratio of the gallery set. Figure \ref{fig:enrollment-setup} provides an overview of our setup. Table \ref{tab:stats} shows the number of identities and walks per identity for each dataset and considered scenarios. We create the training and testing sets for each dataset and scenario using embeddings from pretrained gait recognition models \cite{catruna2024gaitpt,cosma2022learning,teepe2021gaitgraph}.

\begin{table*}[hbt!]
    \centering
     \resizebox{1.0\linewidth}{!}{
    \begin{tabular}{llrrrp{1cm}p{1cm}p{2cm}p{2cm}}
    \textbf{Dataset} & \textbf{Split} & \textbf{ID:Walk Ratio} & \textbf{Gallery Walks} & \textbf{Probe Walks} & \textbf{Unique Gallery IDs} & \textbf{Unique Probe IDs} & \textbf{Walks from new IDs ($y = 1$)} & \textbf{Walks from existing IDs ($y = 0$)} \\
    \midrule
    \multirow{6}{*}{PsyMo} & Train & 0.25 & 370 & 1643 & 31 & 42 & 1115 & 528 \\
     & Test & 0.25 & 168 & 744 & 14 & 19 & 504 & 240 \\
     & Train & 0.5 & 502 & 1511 & 21 & 42 & 503 & 1008 \\
     & Test & 0.5 & 216 & 696 & 9 & 19 & 216 & 480 \\
     & Train & 0.75 & 359 & 1654 & 10 & 42 & 120 & 1534 \\
     & Test & 0.75 & 144 & 768 & 4 & 19 & 48 & 720 \\
     \midrule
     \multirow{6}{*}{CASIA-B} & Train & 0.25 & 702 & 3146 & 26 & 35 & 2156 & 990 \\
     & Test & 0.25 & 297 & 1353 & 11 & 15 & 913 & 440 \\
     & Train & 0.5 & 934 & 2914 & 17 & 35 & 934 & 1980 \\
     & Test & 0.5 & 385 & 1265 & 7 & 15 & 385 & 880 \\
     & Train & 0.75 & 656 & 3192 & 8 & 35 & 224 & 2968 \\
     & Test & 0.75 & 246 & 1404 & 3 & 15 & 84 & 1320 \\
    \end{tabular}
    }
    \caption{Database statistics for training and evaluation scenarios for CASIA-B \cite{yu2006framework} and PsyMo \cite{cosma2023psymo}.}
    \label{tab:stats}
\end{table*}

\subsection{SetTransformer for Gait Enrollment}

\begin{figure}
    \centering
    \includesvg[width=1.0\linewidth]{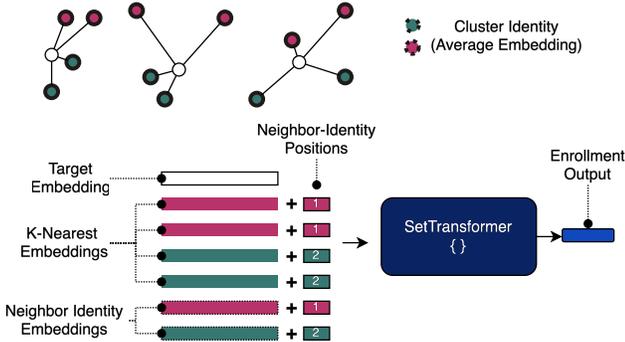}
    \caption{Overall diagram of our model. A probe embedding attends to its $K$ nearest neighbor gait embeddings and corresponding identity‐average embeddings via a SetTransformer to predict enrollment status. The correspondence between identities is facilitated by learned positional embeddings.}
    \label{fig:model-diagram}
\end{figure}

Given a gait embedding, our model's task is to determine whether it belongs to a known identity, already enrolled in the gallery, or a new, unknown identity that should be enrolled. We frame this task as a binary classification problem, where the model uses the input target embedding and relevant samples from the gallery to predict whether the sample is enrolled or not. Because of its training and input, the model is agnostic of the database - it predicts enrollment status only based on the input samples from the gallery. Formally, the gallery set $G$ contains gait representations obtained from a pretrained gait recognition model \cite{catruna2024gaitpt,cosma2022learning,teepe2021gaitgraph}, each gallery embedding $g_k$ having an associated identity $y_k \in \text{ID}$ from a fixed set of identities: $G = \{(g_k, y_k)\}$. The enrollment task implies having a function $f(p, G)$ that decides whether a probe embedding $p$ is part of the gallery identities or not: $f(p, G)\in \{ID, \emptyset\}$.

In practice, we train a neural network on varying configurations of gallery sets for the enrollment task. Since using the whole gallery set $G$ as a model input is not feasible due to its potentially very large scale, we summarize the gallery with a fixed set $K$ of nearest neighbors based on the distance to the probe embedding. As such, given the probe embedding $p$, we compute $\hat{G}_K = \{ (g_i, y_i)|i \leq K\} \subset G \text{;} d(p, g_i) \le d(p, g_{i+1}) \text{ for all } i$.  Furthermore, for each neighbor, we compute an identity embedding by averaging all gallery embeddings that belong to the same identity. Specifically, for a gallery embedding \( g_k \), we define its identity embedding as: $\text{id}_{g_k} = \frac{1}{|\mathcal{G}_{g_k}|} \sum_{g_i \in \mathcal{G}_{g_k}} g_i$, where \( \mathcal{G}_{g_k} = \{ g_j \mid \text{id}_{g_j} = \text{id}_{g_k} \} \) is the set of all gallery embeddings sharing the same identity as \( g_k \).

As such, the input to our model is comprised of a set of the probe embedding $p$, its nearest $K$ neighbors from the gallery $g_k$ and a list of identity embeddings $\text{id}_{g_k}$: $f(p, g_k, \text{id}_{g_k})$. In Figure \ref{fig:model-diagram} we show a diagram of our method. In this work, we propose a model based on the SetTransformer \cite{lee2019set} architecture, leveraging self-attention between the target embedding and selected gallery embeddings, effectively summarizing the most important information in the database concerning the current embedding. We chose to use a SetTransformer since the order of embeddings does not matter in the final enrollment decision, and such approaches have been used in the past for multi-modal processing where the order of modalities is irrelevant \cite{gimeno2024videodepression}. We utilize a two-layer MLP to predict the enrollment status, taking the feature vector predicted by the transformer at the position corresponding to the probe embedding as input. We introduce augmentations such as dropout and random noise to the input embeddings during training.

\noindent \textbf{Modeling neighbor-identity correspondence.}
To effectively incorporate the identity information in the nearest gallery embeddings, we experiment with three different ways to associate the mean identity vector to the corresponding gallery embedding: \textit{Additive Pairing}, \textit{Per-Instance Positions}, and \textit{Per-Identity Positions}. Explicitly associating the identity to each retrieved gallery embedding has the purpose of providing the model with contextual information regarding the state of the gallery and its relation with the current embedding. 

\textbf{Additive Pairing.} The first method implies simply adding the identity information to the corresponding neighbor embeddings. Consequently, we combine each $K$ embeddings with their corresponding identity embedding by computing the element-wise sum. The resulting embeddings are given as input to the model along with the target embedding. This method can be formulated as:

\begin{align*}
    \tilde{g_k} = g_k + id_{g_k}, \quad g_k \in \hat{G}_K, 
\end{align*}

\textbf{Per-Instance Positions.} The second method implies augmenting each neighbor embedding with a unique, learnable positional encoding $P_k$. For each neighbor embedding, the same positional encoding is also added to its corresponding identity embedding. In this method, the model receives both the neighbor embeddings and the identity embeddings. This second method can be formulated as:

\begin{align*}
    \tilde{g_k} &= g_k + P_k \\
    \tilde{id}_{g_k} &= id_{g_k} + P_k, \quad g_k \in \hat{G}_K, 
\end{align*}

\textbf{Per-Identity Positions.} The third method builds upon the second approach by enforcing an additional constraint: neighbor embeddings having the same identity are augmented with the same positional encoding. Doing this encourages the model to focus on identity consistency across neighbors.

\begin{align*}
    \tilde{g_k} &= g_k + P_{id_{g_k}} \\
    \tilde{id}_{g_k} &= id_{g_k} + P_{id_{g_k}}, \quad g_k \in \hat{G}_K, 
\end{align*}

\subsection{Performance Metrics}

To evaluate the performance of our proposed enrollment method, we utilize 4 complementary performance metrics: Matthews Correlation Coefficient \cite{mcc1975}, ROC-AUC, F1 Score, and Average Precision.

\textbf{Matthews Correlation Coefficient (MCC)} is a balanced binary‐classification score with values in the range $[-1,1]$. It is defined as:

\begin{footnotesize}
\begin{align*}
\mathrm{MCC} &= \frac{TP \times TN - FP \times FN}
                {\sqrt{(TP+FP)(TP+FN)(TN+FP)(TN+FN)}}
\end{align*}
\end{footnotesize}

MCC incorporates all four confusion matrix entries and yields +1 for perfect prediction, 0 for random performance, and –1 for total disagreement. MCC is a more appropriate metric for heavily imbalanced classification.

\textbf{Area Under the ROC Curve (ROC–AUC)} is a threshold‐independent measure of binary discrimination defined in the $[0,1]$ interval. It is computed as:
\begin{align*}
\mathrm{AUC} &= \int_0^1 \mathrm{TPR}(t)\,\mathrm{d}\bigl(\mathrm{FPR}(t)\bigr)
\end{align*}
AUC yields +1 for perfect separation, 0.5 for random guessing, and 0 for perfectly inverted predictions.

\textbf{F1 Score} is the harmonic mean of precision ($P$) and recall ($R$) with values ranging from 0 to 1. It is defined as:
\begin{align*}
F_1 &= 2\,\frac{\mathrm{P}\times \mathrm{R}}{\mathrm{P} + \mathrm{R}}
\end{align*}

$F_1$ reaches 1 for perfect precision and recall and 0 when either is zero.

\textbf{Average Precision (AP)} represents the area under the precision–recall curve in $[0,1]$, defined as:

\begin{align*}
\mathrm{AP} &= \int_0^1 \mathrm{Precision}(r)\,\mathrm{d}r
\end{align*}

AP equals 1 for a perfect precision-recall curve and zero if no true positives are retrieved.

\subsection{Datasets Description}
We train the gait embedding models and test our proposed enrollment approach on two gait analysis datasets obtained in controlled laboratory settings: CASIA-B \cite{yu2006framework} and PsyMo \cite{cosma2023psymo}.

CASIA-B \cite{yu2006framework} is a widely utilized gait recognition benchmark consisting of walking sequences of 124 unique individuals recorded from 11 angles. Each participant walks in three different scenarios: normal walking, walking while carrying a bag, and walking while wearing a coat that obstructs parts of the body. Following the most popular data partitioning \cite{catruna2024gaitpt,teepe2021gaitgraph}, we utilize the first 74 subjects for training while keeping the remaining 50 for testing.

PsyMo \cite{cosma2023psymo} is a recently released dataset for analyzing how the walking style of individuals correlates with psychological characteristics. Similarly to CASIA-B, the dataset is collected in controlled settings from multiple viewpoints and scenarios. PsyMo consists of videos of 312 unique subjects recorded from 6 angles across seven scenarios: normal walking, walking while wearing a coat, walking while carrying a bag, walking faster than usual, walking slower than usual, walking while texting, and walking while talking on the phone. The participants were asked to complete several psychological questionnaires, which enabled the exploration of potential links between psychological attributes and walking patterns. However, we only use PsyMo as a gait recognition benchmark and discard psychological annotations. On this dataset, we train the models on the official partitioning in which the first 250 individuals are assigned to the training set while the rest are utilized for evaluation.

\begin{table}[hbt!]
    \centering
     \resizebox{1.0\linewidth}{!}{
    \begin{tabular}{ll|rr}
   \textbf{Dataset} & \textbf{Model} & \textbf{Emb. Size} & \textbf{Accuracy} \\
    \midrule
    \multirow{9}{*}{\textbf{CASIA-B}} & \multirow{3}{*}{\textbf{GaitPT}} & 128 & 0.643 \\
     &  & 256 & 0.645 \\
     &  & 512 & 0.613 \\
     \cmidrule{2-4}
     & \multirow{3}{*}{\textbf{GaitFormer}} & 128 & 0.354 \\
     &  & 256 & 0.463 \\
     &  & 512 & 0.489 \\
     \cmidrule{2-4}
     & \multirow{3}{*}{\textbf{GaitGraph}} & 128 & 0.423 \\
     &  & 256 & 0.433 \\
     &  & 512 & 0.437 \\
     \midrule
    \multirow{9}{*}{\textbf{PsyMo}}& \multirow{3}{*}{\textbf{GaitPT}} & 128 & 0.537 \\
     &  & 256 & 0.532 \\
     &  & 512 & 0.511 \\
     \cmidrule{2-4}
     & \multirow{3}{*}{\textbf{GaitFormer}} & 128 & 0.183 \\
     &  & 256 & 0.314 \\
     &  & 512 & 0.157 \\
     \cmidrule{2-4}
     & \multirow{3}{*}{\textbf{GaitGraph}} & 128 & 0.304 \\
     &  & 256 & 0.311 \\
     &  & 512 & 0.299 \\
    \end{tabular}
    }
    \caption{Recognition accuracy on CASIA-B and PsyMo for three gait‐embedding models (GaitPT, GaitFormer, GaitGraph) for embedding sizes 128, 256, and 512. GaitPT consistently outperforms other models on both datasets.}
    \label{tab:recognition-results}
\end{table}

\subsection{Description of Gait Recognition Models}

In our experiments, to construct gallery and probe sets for use in enrollment, we employ three widely utilized skeleton-based gait recognition models: \textit{GaitGraph} \cite{teepe2021gaitgraph}, \textit{GaitFormer} \cite{cosma2022learning}, and \textit{GaitPT} \cite{catruna2024gaitpt}. All models operate on sequences of pose estimation skeletons and output a vector containing the most relevant movement information for identifying the person, but they differ in their underlying architectures. Table \ref{tab:recognition-results} showcases gait recognition results for all three models under different embedding sizes.

\textbf{GaitPT \cite{catruna2024gaitpt}} is a hierarchical transformer model with good results in skeleton-based gait recognition. The architecture employs a hierarchical pipeline of four stages, incorporating spatio-temporal attention at multiple structural levels: joint, limb, limb group, and body. This approach enables it to capture discriminative walking features at all levels of movement.

\textbf{GaitGraph \cite{teepe2021gaitgraph}.} utilizes a residual graph convolution (ResGCN \cite{song2020stronger}) architecture to extract discriminative spatial and temporal features from the walking sequence in order to identify subjects.

\textbf{GaitFormer \cite{cosma2022learning}} is among the first attention-based approaches to gait recognition and adopts a transformer-encoder \cite{vaswani2017attention} model, which computes self-attention at the skeleton level, effectively modeling the evolution of movement during walking. Since each skeleton is handled as a separate token, the model does not explicitly handle spatial relationships between joints.

We train all architectures on sequences of 48 consecutive skeletons using triplet loss, which aims to minimize the embedding distance of walks belonging to the same identity while maximizing the distance between those belonging to different subjects. A nearest-neighbor approach is utilized to identify a new walking sample, comparing its embedding with embeddings of known walking sequences. For evaluation, we follow each dataset's evaluation protocol and compute the average performance across viewpoints and scales, excluding identical view cases. Each model is trained using three different embedding sizes: 128, 256, and 512.

\section{Results}
\label{sec:results}
\begin{figure}[hbt!]
    \centering
    \includesvg[width=1.0\linewidth]{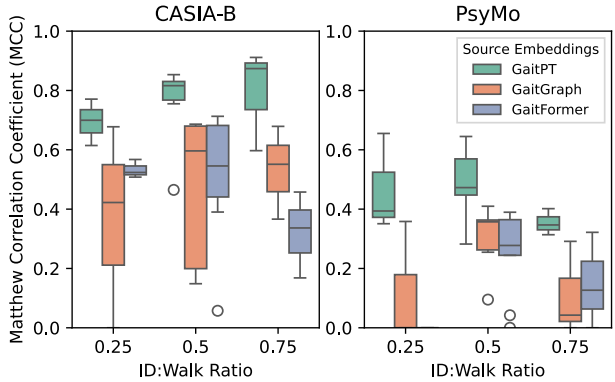}
    \caption{Performance across ID: Walk ratios (0.25, 0.5, 0.75) on CASIA-B and PsyMo for GaitPT, GaitGraph, and GaitFormer. On average, the more imbalanced the setup, the more performance drops.}
    \label{fig:id-ratio}
\end{figure}

\noindent \textbf{Evaluation on different enrollment scenarios.} Enrollment performance hinges on the ratio of identities to samples per identity. As Figure \ref{fig:id-ratio} shows, having fewer identities with more walks each leads to significantly higher MCC than the opposite case of many identities with few samples. However, increasing data imbalance results in a drop in performance for almost all models. The model using GaitPT embeddings obtains the best performance, consistent with its high results in gait recognition compared to the other models (Table \ref{tab:recognition-results}).

\begin{figure}[hbt!]
    \centering
    \includesvg[width=1.0\linewidth]{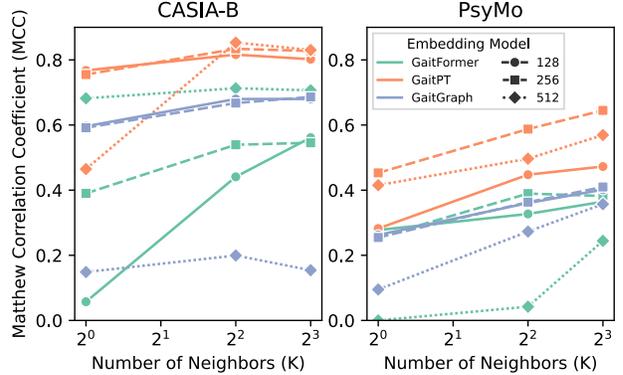}
    \caption{The effect of increasing the number of neighbors $K$ in the SetTransformer, on CASIA-B and PsyMo, for GaitPT, GaitGraph and GaitFormer, with embedding sizes 128, 256, 512. More neighbors present in the context results in better enrollment performance.}
    \label{fig:k-ablation}
\end{figure}

\noindent \textbf{Effect of number of neighbors $K$.} The number of nearest neighbors \(K\) has a major impact on enrollment. As Figure \ref{fig:k-ablation} shows, MCC rises steadily with larger \(K\) across all embedding types and sizes. This demonstrates that drawing on a wider neighborhood strengthens the model’s decisions. At the limit, with $K$ equaling the dataset size, the database configuration is fully determined. As such, larger $K$ provides more context to the full database.

\begin{figure}[hbt!]
    \centering
    \includesvg[width=1.0\linewidth]{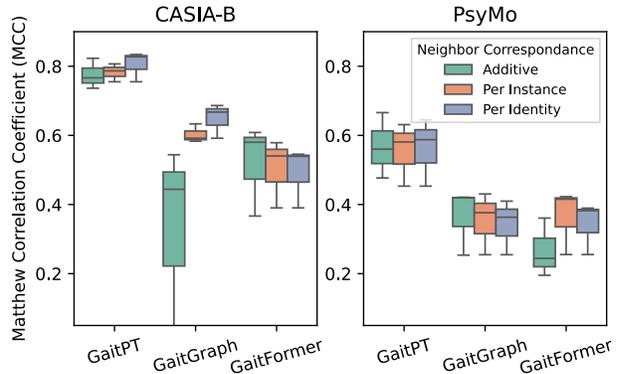}
    \caption{Comparison of neighbor–identity coupling methods (Additive, Per-Instance, Per-Identity) in terms of MCC \cite{mcc1975} on CASIA-B and PsyMo using GaitPT, GaitGraph and GaitFormer generated embeddings.}
    \label{fig:neighbor-type-ablation}
\end{figure}

\noindent \textbf{Effect of the neighbor-identity correspondence.} Providing explicit identity context when combining neighbors with their identity means boosts performance. In Figure \ref{fig:neighbor-type-ablation}, the Per-Identity Positions approach—where all examples of an identity share the same learnable position code achieves the highest median MCC and lowest variance on both CASIA-B and PsyMo. This shows that linking neighbors by identity helps the SetTransformer attend to true identity groups rather than isolated samples.

\begin{table*}[hbt!]
    \centering
    \resizebox{0.75\linewidth}{!}{
    \begin{tabular}{lr|rrrr}
    \textbf{Model} & \textbf{ID:Walk Ratio} & \textbf{F1} & \textbf{ROC-AUC} & \textbf{AP} & \textbf{MCC} \\
    \midrule
    PCA + Logistic Reg. on Embeddings &  \multirow{3}{*}{0.25} & 0.664 & 0.582 & 0.714 & 0.154 \\
    Logistic Reg. on Embeddings &  & 0.844 & 0.672 & 0.761 & 0.427 \\
    Logistic Reg. on Neighbour Distances &  & \textbf{0.939} & \textbf{0.941} & \textbf{0.961} & \textbf{0.844} \\
    \textbf{SetTransformer (ours)} &  & 0.865 & 0.817 & 0.859 & 0.615 \\
    \midrule
    PCA + Logistic Reg. on Embeddings & \multirow{3}{*}{0.5} & 0.466 & 0.586 & 0.350 & 0.158 \\
    Logistic Reg. on Embeddings & & 0.573 & 0.682 & 0.415 & 0.338 \\
    Logistic Reg. on Neighbour Distances &  & 0.866 & 0.882 & \textbf{0.836} & \textbf{0.832} \\
    \textbf{SetTransformer (ours)} & & \textbf{0.881} & \textbf{0.910} & 0.818 & 0.830 \\
    \midrule
    PCA + Logistic Reg. on Embeddings & \multirow{4}{*}{0.75} & 0.071 & 0.363 & 0.050 & -0.145 \\
    Logistic Reg. on Embeddings &  & 0.081 & 0.499 & 0.060 & -0.002 \\
    Logistic Reg. on Neighbour Distances & & 0.756 & 0.804 & 0.631 & 0.770 \\
    \textbf{SetTransformer (ours)} & & \textbf{0.917} & \textbf{0.956} & \textbf{0.845} & \textbf{0.911} \\
    \end{tabular}
    }
    \caption{Comparisons with baseline methods for enrollment for CASIA-B. Our method obtains better results especially in scenarios where there are many walks per identity.}
    \label{tab:baseline}
\end{table*}

\begin{table*}[hbt!]
    \centering
     \resizebox{0.75\linewidth}{!}{
    \begin{tabular}{lr@{$\longrightarrow$}l|rrrr}
    \textbf{Source Emb.} & \multicolumn{2}{c|}{\textbf{Train Dataset} $\longrightarrow$ \textbf{Test Dataset}}\ & \textbf{F1} & \textbf{ROC-AUC} & \textbf{AP} & \textbf{MCC} \\
    \midrule
    GaitFormer & CASIA-B & PsyMo & 0.492 & 0.625 & 0.390 & 0.242 \\
    GaitFormer & PsyMo & CASIA-B & 0.333 & 0.578 & 0.366 & 0.216 \\
    \midrule
    GaitPT & CASIA-B & PsyMo & 0.421 & 0.571 & 0.349 & 0.138 \\
    GaitPT & PsyMo & CASIA-B & 0.378 & 0.555 & 0.334 & 0.112 \\
    \midrule
    GaitGraph & CASIA-B & PsyMo & 0.498 & 0.547 & 0.332 & 0.178 \\
    GaitGraph & PsyMo & CASIA-B & 0.464 & 0.499 & 0.303 & -0.007 \\
    \end{tabular}
    }
    \caption{Cross‐embedding enrollment performance on CASIA-B and PsyMo. For each pairing of training and testing embeddings, this table reports F1, ROC–AUC, AP and MCC.}
    \label{tab:crossval-dataset}
\end{table*}

\begin{figure}[hbt!]
    \centering
    \includesvg[width=0.85\linewidth]{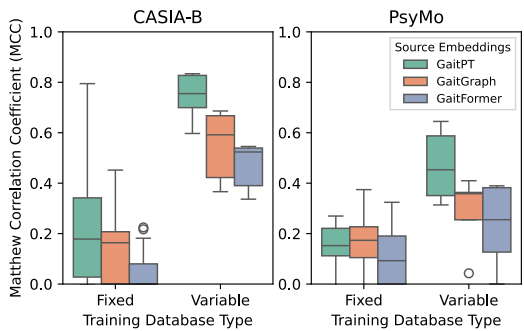}
    \caption{Enrollment MCC on CASIA-B and PsyMo for fixed vs. variable training database types using GaitPT, GaitGraph and GaitFormer. Training with variable database types significantly improves performance, as the model does not overfit a particular configuration.}
    \label{fig:db-type}
\end{figure}

\noindent \textbf{Effect of varying the training database.} Enrollment robustness depends on whether the training database is held fixed or allowed to vary. As Figure \ref{fig:db-type} illustrates, switching to a variable database setup — in which different identity subsets and sampling distributions are seen during training — yields uniformly higher MCC on both CASIA-B and PsyMo, regardless of the source embedding. GaitPT, GaitGraph, and GaitFormer all benefit from this diversity, with GaitPT showing the largest absolute increase. These results suggest that exposing the enrollment model to a broader range of gallery configurations during training enhances its ability to generalize to unseen identity distributions.

\noindent \textbf{Comparison with baseline models.} In Table \ref{tab:baseline}, we compare the SetTransformer with three methods for doing enrollment: we train a Logistic Regression model that operates directly on neighbor distances and outputs the enrollment label, similar to Platt scaling \cite{platt2000scaling}, a Logistic Regression model that operates on the concatenated probe and associated neighbor embeddings (resulting in a large 512 * 8 = 4096 feature vector) and similarly, a Logistic Regression that operates on embeddings compressed with a PCA model to avoid high dimensionality. Results are shown for CASIA-B, using GaitPT 512-dimensional embeddings and $K = 8$. Our model obtains the overall best results in the scenario with ID:Walk ratio of 0.75, which contains the most training embeddings. This highlights the fact that using SetTransformers is a flexible method for performing post-training enrollment that scales better with data that traditional methods.

\begin{table*}[hbt!]
    \centering
     \resizebox{0.75\linewidth}{!}{
        \begin{tabular}{l r@{$\longrightarrow$}l | r r r r}
          \textbf{Dataset}
            & \multicolumn{2}{c|}{\textbf{Train Emb. $\longrightarrow$ Test Emb.}}
            & \textbf{F1}
            & \textbf{ROC-AUC}
            & \textbf{AP}
            & \textbf{MCC} \\
          \midrule
          \multirow{6}{*}{\textbf{CASIA-B}}
            & GaitFormer & GaitPT    & 0.228 & 0.560 & 0.373 & 0.256 \\
            & GaitFormer & GaitGraph & 0.179 & 0.549 & 0.370 & 0.260 \\
            & GaitPT      & GaitFormer & 0.571 & 0.693 & 0.468 & 0.400 \\
            & GaitPT      & GaitGraph & 0.479 & 0.544 & 0.324 & 0.109 \\
            & GaitGraph   & GaitFormer & 0.328 & 0.578 & 0.367 & 0.221 \\
            & GaitGraph   & GaitPT    & 0.430 & 0.609 & 0.383 & 0.245 \\
          \midrule
          \multirow{6}{*}{\textbf{PsyMo}}
            & GaitFormer & GaitPT    & 0.412 & 0.512 & 0.316 & 0.023 \\
            & GaitFormer & GaitGraph & 0.453 & 0.563 & 0.342 & 0.116 \\
            & GaitPT      & GaitFormer & 0.479 & 0.511 & 0.315 & 0.085 \\
            & GaitPT      & GaitGraph & 0.528 & 0.620 & 0.375 & 0.235 \\
            & GaitGraph   & GaitFormer & 0.474 & 0.500 & 0.310 & 0.000 \\
            & GaitGraph   & GaitPT    & 0.474 & 0.501 & 0.311 & 0.025 \\
        \end{tabular}
      }
    \caption{Cross‐embedding enrollment performance on CASIA-B and PsyMo. Each row reports F1, ROC–AUC, AP and MCC when training embeddings are from one gait model (Train Emb.) and the testing in the enrollment task is from another (Test Emb.).}
    \label{tab:crossval-model}
\end{table*}
\noindent \textbf{Cross-dataset evaluation.} Evaluations across datasets (Table \ref{tab:crossval-dataset}) show a clear domain gap: models trained on CASIA-B lose accuracy when tested on PsyMo embeddings, and vice versa. GaitFormer embeddings hold up best when moving from CASIA-B to PsyMo (ROC-AUC 0.625, F1 0.492), while GaitGraph ones nearly collapse in the opposite direction (MCC of 0.007). GaitPT sits in the middle, with MCCs around 0.11-0.14 for both transfer directions. For cross-dataset enrollment, embeddings from GaitFormer seem to generalize best. 

\noindent \textbf{Cross-model evaluation.} Evaluations across model embeddings (Table \ref{tab:crossval-model}), show that different embedding backbones prove complementary. On CASIA-B, a classifier trained on GaitPT embeddings and tested on GaitFormer embeddings achieves the best scores (MCC 0.400, ROC–AUC 0.693), whereas any scenario involving GaitGraph embeddings generally underperforms (e.g., GaitFormer $\longrightarrow$ GaitGraph yields an F1 of 0.179). For PsyMo, the highest MCC (0.235) comes from training on GaitPT embeddings and testing on GaitGraph embeddings. 

\section{Discussion and Conclusions}
\label{sec:conclusions}
This work tackles the problem of enrollment in the context of open-set gait recognition. We propose a setup designed to automatically generate diverse training data with various enrollment scenarios from existing gait recognition datasets. Utilizing these enrollment scenarios, we train an attention-based model to predict whether new gait samples belong to known identities or should be enrolled in the gallery. Our model operates on embeddings obtained with widely utilized gait recognition architectures and makes the prediction based on the contextual information from the dataset in the form of relevant embeddings.

Our proposed method is both: \textit{(i)} dataset agnostic as it learns to predict enrollment status based on the context of the dataset and observes various gallery sets during training, and \textit{(ii)} model agnostic as it operates directly on the embeddings produced by a recognition architecture and does not observe the input modality.

We evaluate our approach on two widely utilized benchmarks for gait recognition in laboratory settings (CASIA-B \cite{yu2006framework} and PsyMo \cite{cosma2023psymo}) and employ three skeleton-based identification architectures (GaitPT \cite{catruna2024gaitpt}, GaitFormer \cite{cosma2022learning}, and GaitGraph \cite{teepe2021gaitgraph}) as part of our analysis on gait enrollment. The results show that our approach to gait enrollment generalizes across different settings of identities, samples per identities, gait recognition architectures, and datasets. This work represents a step toward practical deployments of gait recognition systems capable of deciding between identification and enrollment.

{\small
\bibliographystyle{ieee}
\bibliography{refs}
}

\end{document}